\documentclass[runningheads]{llncs}

\usepackage{eccv}

\usepackage{eccvabbrv}

\usepackage{graphicx}
\usepackage{booktabs}

\usepackage[accsupp]{axessibility}  

\usepackage{hyperref}

\usepackage{orcidlink}

\begin{document}

\title{Latent Distillation for Continual Object Detection at the Edge} 

\titlerunning{Latent Distillation}

\author{Francesco Pasti\inst{1}\orcidlink{0009-0005-4992-3281} \and
Marina Ceccon\inst{1}\orcidlink{0009-0006-5892-4080} \and
Davide Dalle Pezze\inst{1}\orcidlink{0000-0002-4741-1021} \and
Francesco Paissan\inst{2}\orcidlink{0000-0002-5553-7935} \and
Elisabetta Farella\inst{2}\orcidlink{0000-0001-9047-9868} \and
Gian Antonio Susto\inst{1}\orcidlink{0000-0001-5739-9639}\and
Nicola Bellotto\inst{1}\orcidlink{0000-0001-7950-9608}}

\authorrunning{Pasti et al.}

\institute{ University of Padova, Italy\\
\email{francesco.pasti@dei.unipd.it, marina.ceccon@phd.unipd.it, {\{davide.dallepezze, gianantonio.susto, nicola.bellotto\}@unipd.it}}\\ \and
Fondazione Bruno Kessler, Italy\\
\email{\{fpaissan, efarella\}@fbk.it}}

\maketitle

\begin{abstract}
While numerous methods achieving remarkable performance exist in the Object Detection literature, addressing data distribution shifts remains challenging. Continual Learning (CL) offers solutions to this issue, enabling models to adapt to new data while maintaining performance on previous data. This is particularly pertinent for edge devices, common in dynamic environments like automotive and robotics.
In this work, we address the memory and computation constraints of edge devices in the Continual Learning for Object Detection (CLOD) scenario.
Specifically, (i) we investigate the suitability of an open-source, lightweight, and fast detector, namely NanoDet, for CLOD on edge devices, improving upon larger architectures used in the literature. Moreover, (ii) we propose a novel CL method, called Latent Distillation~(LD), that reduces the number of operations and the memory required by state-of-the-art CL approaches without significantly compromising detection performance. Our approach is validated using the well-known VOC and COCO benchmarks, reducing the distillation parameter overhead by 74\% and the Floating Points Operations~(FLOPs) by 56\% per model update compared to other distillation methods.
\keywords{Continual Learning \and Edge Devices \and Object Detection \and Distillation}
\end{abstract}

\section{Introduction}
\label{sec:intro}
\begin{figure}[tb]
\begin{tabular}{c@{\hspace{5pt}}|@{\hspace{5pt}}c}
\includegraphics[width=0.55\linewidth]{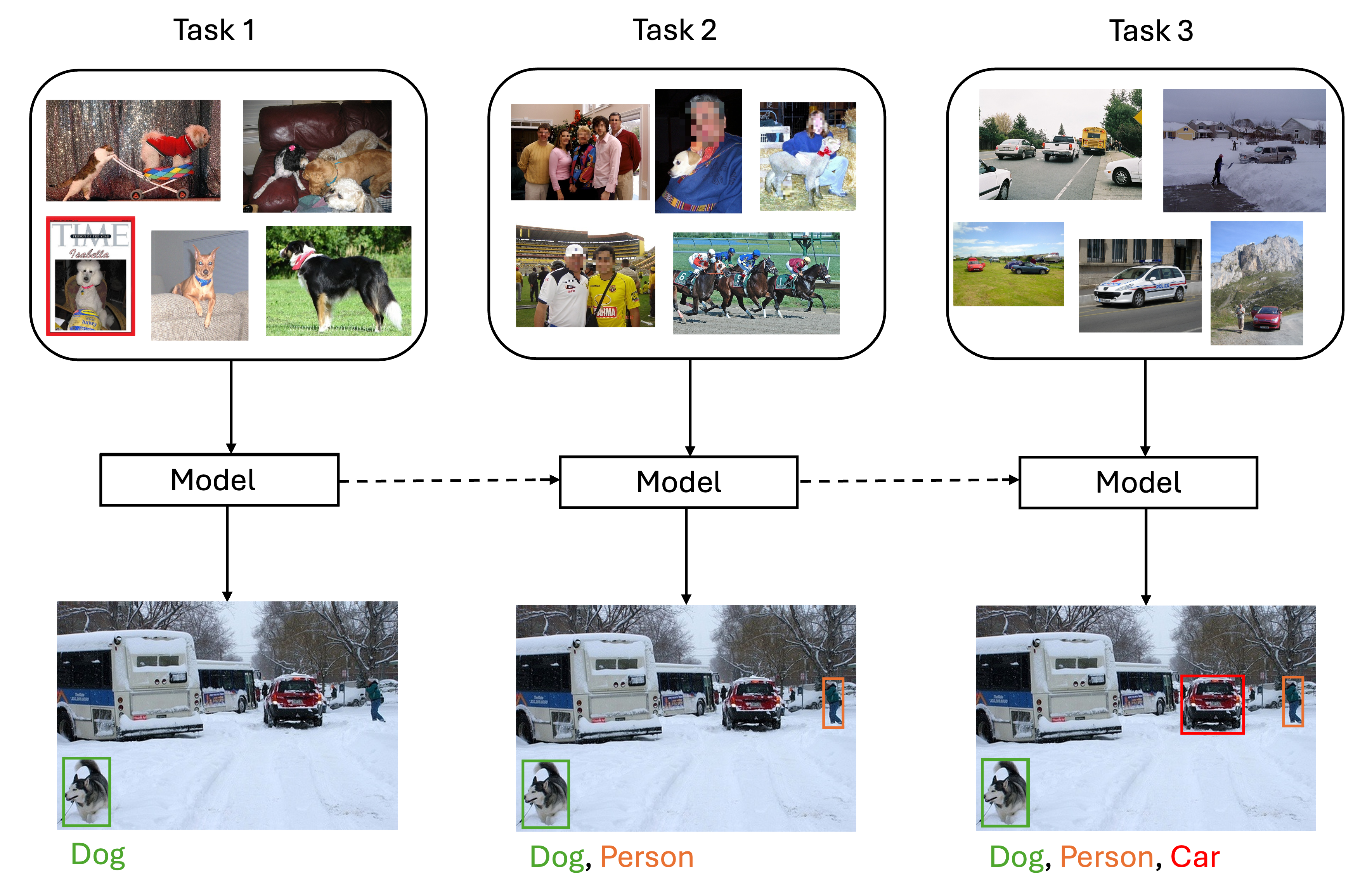}&
\includegraphics[width=0.42\linewidth]{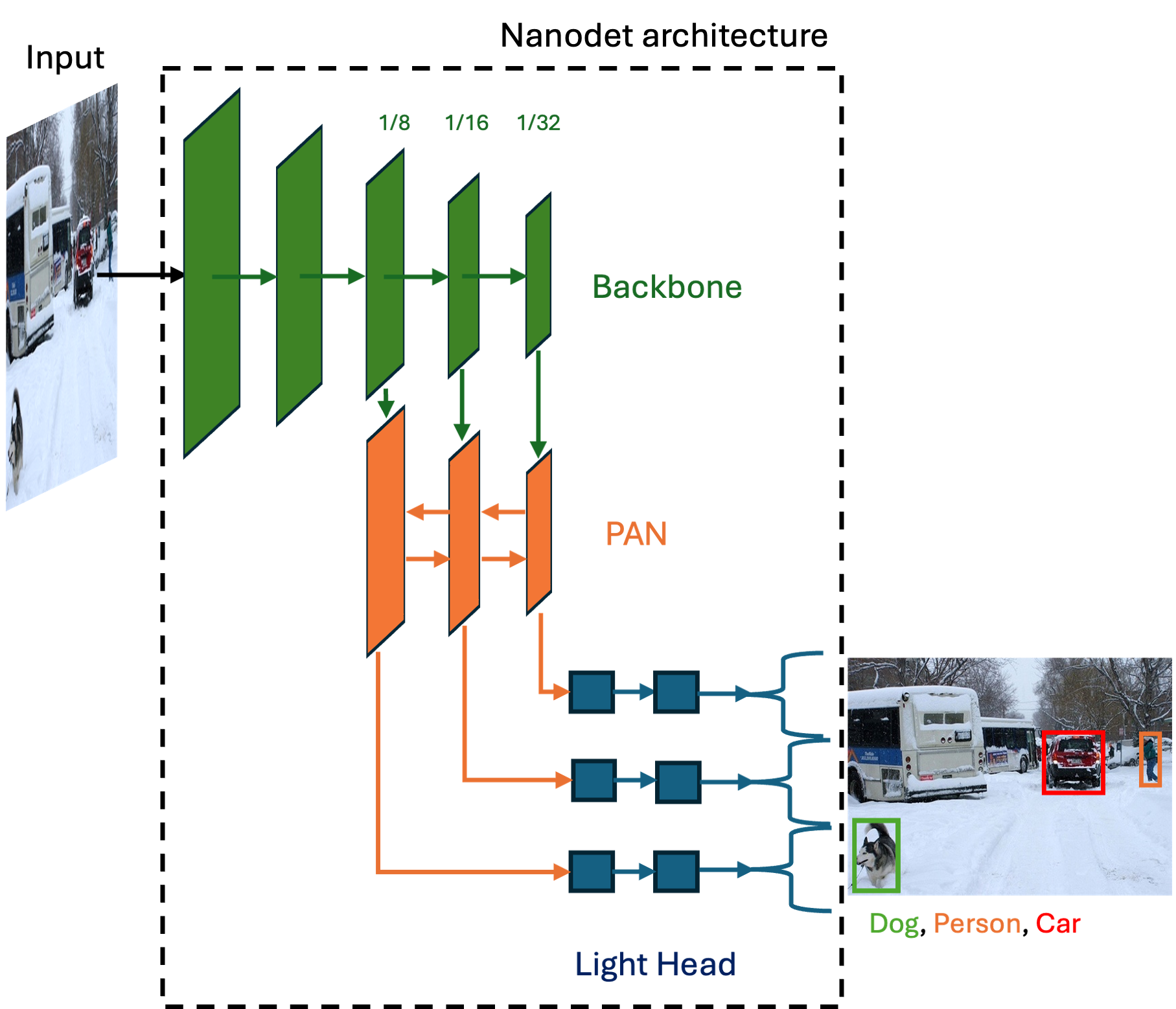}\\
\end{tabular}
\caption{\textbf{Left:} Continual Learning setting for Object Detection. From one task to the next, the model must learn from a new dataset where only new classes are annotated. The model is expected to learn new classes without compromising performance on previous ones as depicted in the bottom row. \textbf{Right:} Open-source NanoDet architecture. The image is fed through the backbone, three of its hidden representations are used by the GhostPAN feature pyramid and later fed to the lightweight detection heads.}
\label{fig:CLOD_scheme}
\end{figure}
Object detection is a computer vision task that involves identifying and locating objects within an image. Its application spans numerous domains, such as robotics, medical imaging, and manufacturing~\cite{karaoguz2019object, li2019clu}. 
Despite significant advancements and impressive performance of recent solutions~\cite{zaidi2022survey}, some challenges still need to be addressed to deploy state-of-the-art algorithms in real-world scenarios.

One of these challenges involves the distribution shifts often present in new data~\cite{guan2018learn}. 
For example, the environment in which a moving agent operates could change, introducing new classes or presenting objects in different contexts~\cite{vandeven2019}.
A common approach to handle these data shifts is to fine-tune the model on the new domain.
However, this leads to the catastrophic forgetting phenomenon where the model entirely forgets the previously acquired knowledge~\cite{kirkpatrick2017overcoming,pezze2022continual}.
Continual Learning (CL) addresses this issue, enabling the model to adapt to new data while retaining knowledge of the old~\cite{hadsell2020embracing}.

Several methods from the Continual Learning literature~\cite{replaypaper, lopez2017gradient, li2017learning} tackle the Class Incremental Learning (CIL) problem.
In this scenario, images with newly labeled classes appear in new tasks. The model must learn to detect the new classes without forgetting the old ones. While most works in the CIL setting focus on image classification~\cite{mai2022online}, less attention is given to object detection which poses additional challenges. Indeed, class instances labeled in old tasks may be present in new task data, even though their information is no longer present in the ground truth.
Additionally, objects previously learned by the model as background, may represent the new class data as depicted in~\cref{fig:CLOD_scheme}~(left).

The CL framework is also valuable in scenarios where storing all the tasks data is too expensive, retraining from scratch is computationally infeasible, and a fast model update is required, such as in edge applications~\cite{ramjattan2023using}.
For example, Continual Learning for Object Detection (CLOD) at the edge is particularly relevant for tiny robots~\cite{neuman2022tiny}.
However, most CLOD approaches in the literature~\cite{menezes2023continual} are still not suitable for real-world scenarios with limited computing resources, especially in edge and tinyML applications with microcontrollers~(MCUs) where both the model size and computational requirements for updates must be minimized.
The contribution of our manuscript is then twofold:

\noindent \textbf{(i)} We investigate and experimentally evaluate the feasibility of an efficient open-source object detector, namely NanoDet~\cite{rangilyu_rangilyunanodet_2024}, for CLOD applications at the edge. We study whether a lightweight model can offer sufficient plasticity on new data while preventing catastrophic forgetting of old data.

\noindent \textbf{(ii)} To further reduce the computational and memory burden required when updating the model in CLOD applications, we propose and benchmark a new method, Latent Distillation~(LD). LD reduces the number of FLOPs and parameters required by state-of-the-art CL approaches with a negligible degradation in detection performance.

Furthermore, we publicly release our code \footnote{\url{https://github.com/pastifra/Continual_Nanodet}} to replicate the results and promote further research in the field.

The paper is structured as follows:
~\cref{sec:related_work} provides an overview of the related works focusing on CLOD for edge applications. 
In~\cref{sec:our_approach}, we describe our approach for CLOD at the edge.~\cref{sec:experimental_setting} introduces the experimental setting, including the benchmark datasets, compared CL methods, and evaluation metrics.
The results of the experiments are then presented in~\cref{sec:results}.
Finally, in~\cref{sec:conclusions_future_works}, we conclude and discuss future research directions.

\section{Related Works}\label{sec:related_work}
Continual Learning allows deep learning models to update and expand their knowledge over time with minimal computation and memory overhead~\cite{de2021continual}. An effective CL solution, should minimize forgetting of old tasks, require low memory, and be computationally efficient during incremental model updates.
As defined in~\cite{vandeven2019}, the literature explores three CL scenarios: Task-Incremental~(TIL), Domain-Incremental~(DIL), and Class-Incremental~(CIL). Most CLOD works~\cite{shmelkov2017incremental,acharya2020rodeo,li2019rilod,peng_sid_2021} focus on the CIL scenario, where images with annotations of new classes are presented in each new task. Images associated with the previous classes may still be included in the task samples, but the old classes are not labeled in the new tasks. Consequently, the model must retain knowledge of the old task classes and accurately assign the previous labels while adapting to the new task.

CL strategies can be grouped into three clusters: (i) rehearsal-based~\cite{lopez2017gradient, pmlr-v162-guo22g}, (ii) regularization-based~\cite{kirkpatrick2017overcoming, li2017learning}, and (iii) architecture-based~\cite{sokar2021spacenet, mallya2018piggyback}. In this work, we explore rehearsal-based and regularization-based approaches, as a recent survey on CLOD~\cite{menezes2023continual} shows they are the most effective strategies to mitigate the effects of catastrophic forgetting in object detection networks.

Rehearsal-based methods require storing and reusing past data samples during training. Within this category, various approaches exist, with one of the most renowned methods being Experience Replay~\cite{rolnick2019experience}. Regularization-based techniques, on the other hand, introduce additional constraints or penalties during training to maintain the memory of old tasks.
For example, they might penalize the parameters differently based on their importance~\cite{kirkpatrick2017overcoming} or use distillation to retain knowledge from previous tasks~\cite{li2017learning}.
In the object detection setting, regularization-based approaches, pioneered by~\cite{shmelkov2017incremental}, typically use a classic distillation teacher-student setup, relying on a Region Proposal Network (RPN) to get accurate bounding box predictions for new classes. Other works~\cite{chen2019new, peng_sid_2021} introduce intermediate distillation to further mitigate catastrophic forgetting.
IncDet~\cite{liu2020incdet}, in contrast to the distillation setup, constraints the parameter update with Elastic Weight Consolidation (EWC). However, it needs to use pseudo-labels and novel Huber regularization loss to achieve good performances in the CLOD scenario.
Replay-based approaches are also typically combined with other CL strategies to achieve good performances for CLOD. A relevant example is~\cite{hao2019take}, which combines a small replay buffer with logits distillation.

CL techniques based on Replay are also demanding in terms of computation and storage. They require storing some images from previous training tasks in a memory buffer and updating parameters for the entire network. Additionally, since each training batch adds samples from the memory buffer, each model update takes longer.
To mitigate these issues, Latent Replay~(LR)~\cite{pellegrini2020latent} freezes the first layers of the network and stores the produced latent activations in the replay memory, reducing both the memory and computation required by Replay.
While this method proved effective for a tinyML platform requiring only 64MB of memory~\cite{ravaglia2021tinyml}, it still relies on a replay memory, whose size could still be too heavy for some edge applications.
Building on these ideas, we propose Latent Distillation~(LD) that, by distilling knowledge on latent layers, minimizes the number of FLOPs and parameters to update the model without the need to store and train on the memory buffer samples.

The time necessary to train the model on a new task and the resources required to do so are two important constraints that must be taken into account when considering CLOD at the edge.
Most of the works proposed in CLOD literature~\cite{hao2019end,peng2020faster,zhou2020lifelong,acharya2020rodeo,yang2022rd} consider two-stage detectors like Faster R-CNN~\cite{ren2015faster} which combines a RPN with a classifier.
However, these solutions are sub-optimal for edge applications as they imply longer inference time and two separate training phases, resulting in a less efficient architecture compared to one-stage detectors.
Few works in CLOD literature consider the constraints of edge applications.
In~\cite{li2019rilod}, the authors use a one-stage detector, RetinaNet~\cite{lin2017focal} with a ResNet-50 backbone.
However, such model is not feasible for deployment on constrained edge devices as it has 34M parameters. Common MCUs used to run tinyML edge applications~\cite{phinet, abadade2023comprehensive, 9746093} typically consider 2M of \texttt{int8} encoded weights as the limit for the model.
Other works~\cite{zhang2021incremental,ul2021incremental,shieh2020continual} use a YOLO architecture~\cite{jiang2022review}, significantly reducing inference time.
Another step towards edge applications was performed by two works~\cite{nenakhov2021continuous, ramjattan2023using} that consider the use of YOLOv5s, a smaller version of YOLOv5 more suitable for edge devices, although still challenging to run on edge devices since it has 7.5M parameters.
In this work, we target a tiny object detection network for CLOD, Nanodet~\cite{rangilyu_rangilyunanodet_2024}, which includes a variant with only 1.2M parameters requiring 0.9G FLOPs for inference.

\section{CLOD at the Edge}
\label{sec:our_approach}
The following section describes the proposed approach for CLOD at the edge. First, in~\cref{subsec:nanodet}, we describe the lightweight open-source detector, NanoDet~\cite{rangilyu_rangilyunanodet_2024}, and how we adapted it for CLOD at the edge. Then, in~\cref{subsec:latent_dist} we provide a comprehensive explanation of the proposed CL technique Latent Distillation.
\subsection{Efficient Object Detection Pipeline}
\label{subsec:nanodet}

\begin{figure}[tb]
\begin{tabular}{c|c}
\includegraphics[width=0.5\linewidth]{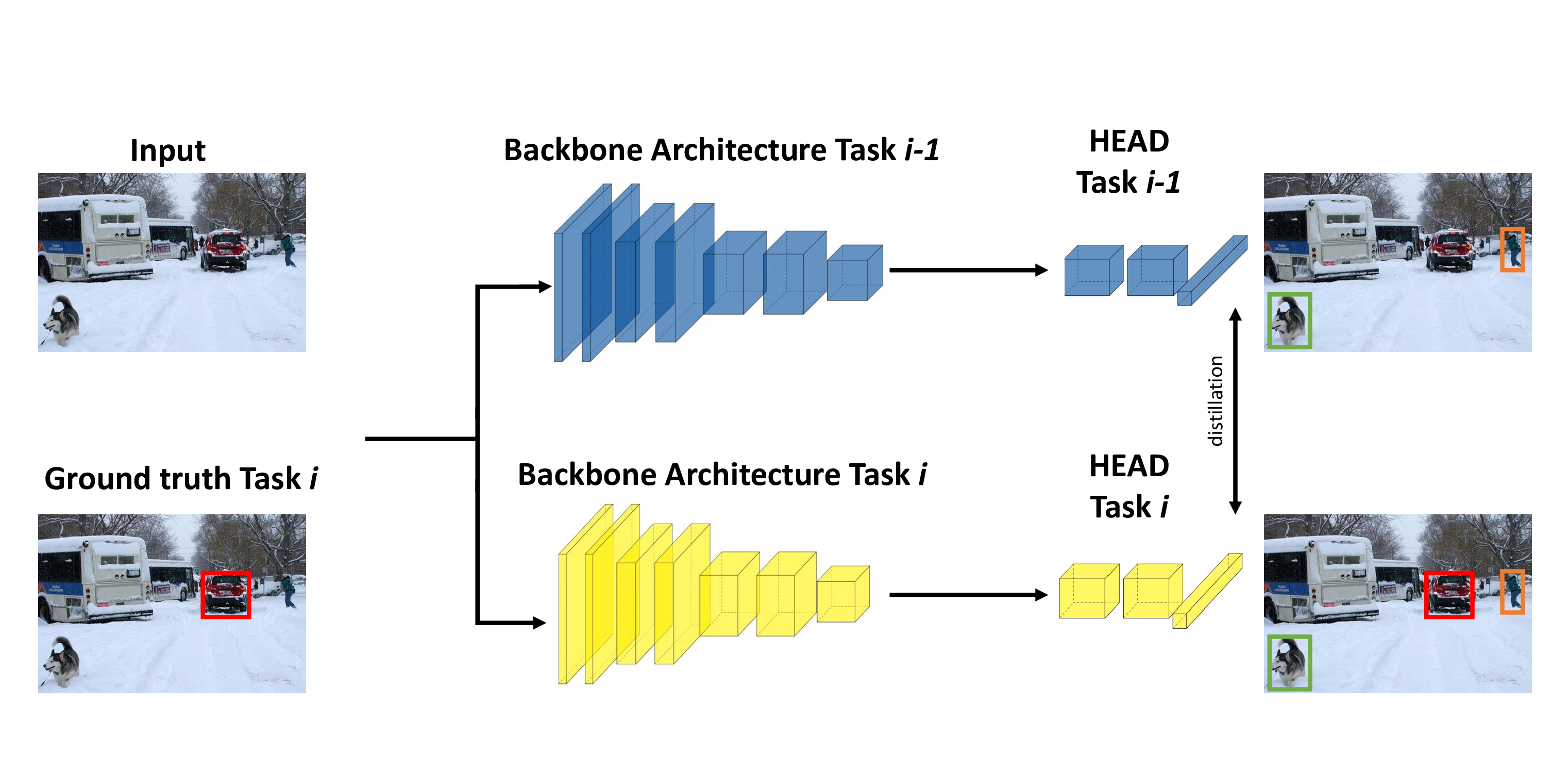}&
\includegraphics[width=0.5\linewidth]{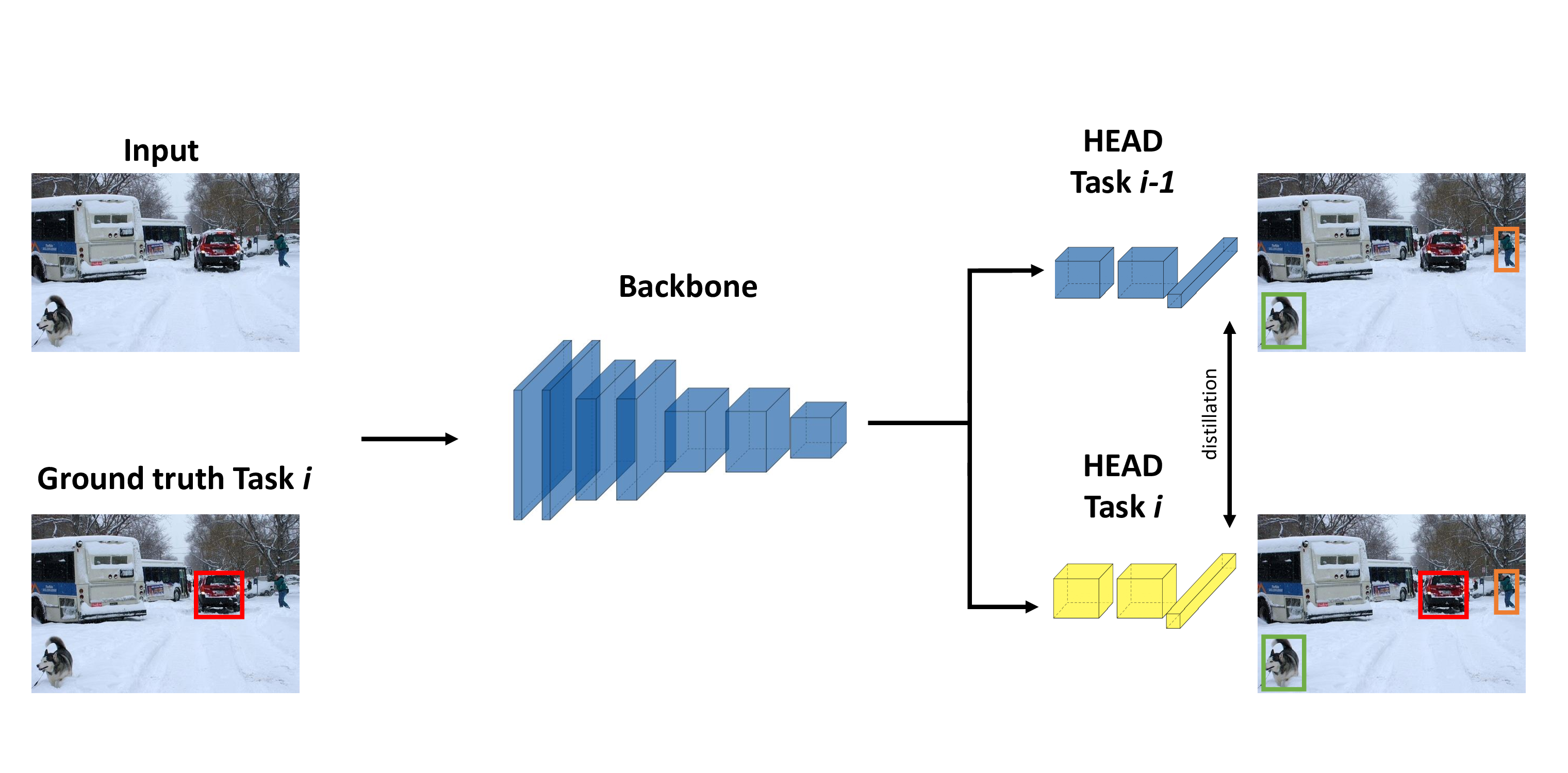}\\
\end{tabular}
\caption{\textbf{Left:} Classic distillation, the teacher is frozen (in blue) and distills knowledge to the student. In this setting, the memory occupation of the CL strategy is double the amount of parameters of the detection network. \textbf{Right:} Latent Distillation, student and teacher share a common frozen part. This enables reducing the memory and computation of the CL strategy.}
\label{fig:OurApproach}
\end{figure}

For CLOD at the edge, we use NanoDet~\cite{rangilyu_rangilyunanodet_2024}, an open-source anchor-free object detector developed for real-time inference on edge devices.
NanoDet falls in the category of Fully Convolutional One-Stage object detectors (FCOS)~\cite{tian2020fcos}. These detectors are composed of (i) a backbone that extracts features representing the objects in the frame; (ii) a feature pyramid neck (FPN) that upscales the representations of the backbone and abstracts them at different scales; and (iii) a detection head, responsible for predicting the bounding boxes and their respective classes.

NanoDet, depicted in~\cref{fig:CLOD_scheme}~(right), presents a good performance-complexity tradeoff.
Its improvement over other detectors can be associated with four characteristics: (i) an efficient edge-oriented backbone, namely ShuffleNetv2~\cite{ma2018shufflenet}; (ii) an efficient feature-pyramid network, named GhostPAN; (iii) the merging of the centerness and classification score thanks to the Generalized Focal Loss~\cite{li2020generalized}; and (iv) the introduction of an auxiliary module to solve the optimal label assignment problem, adding some training overhead.
Moreover, in the CL settings, the optimal anchors for the new task bounding boxes can potentially shift from the ones defined for the old data. As NanoDet is anchor-free, the model regression outputs are free to adapt to the new bounding boxes.

We implement Nanodet-Plus-m from~\cite{rangilyu_rangilyunanodet_2024}, which takes as input 320x320x3 images and uses the ShuffleNetv2 backbone, achieving 27\% of mean Average Precision~(mAP) on COCO.
Despite being a good choice for edge deployment, this architecture considers weight averaging and an auxiliary head to achieve the best performance, therefore it has 4.8M parameters during training.
Since the target of our work is to push the state-of-the-art for efficient CLOD, we reduced the training overhead of NanoDet by removing the auxiliary head and weight averaging modules, decreasing the overall parameters count to 1.2M, with a minor 2\% drop of mAP on COCO, while speeding up the training process.

\subsection{Latent Distillation}\label{subsec:latent_dist}
We identify two main weaknesses when considering classic distillation approaches for the edge.
First, to perform distillation, the teacher model, which is trained on the previous task, introduces a training overhead as its logits need to be computed at each forward step. Moreover, training the entire network is computationally expensive and might not be necessary to retain the old knowledge while adapting to the new tasks.
To address these issues, following the intuition of Latent Replay~\cite{pellegrini2020latent}, we propose a simple yet effective strategy, Latent Distillation (LD).

\begin{algorithm}
\small
\caption{Latent Distillation}\label{alg:LD}
\begin{algorithmic}
\STATE \textbf{Start with:}
\STATE Lower frozen layers $f_\theta$, parameterized by weights $\theta$; Upper Layers $h_\phi$, parameterized by weights $\phi$; Model $G(x;\theta;\phi) = h_{\phi}(f_{\theta}(x))$ of task $n-1$; $c$ is the number of classes until task $n-1$; $k$ is the number of classes to learn at task $n$; Update rule $\mathcal{U}$; Training data ($X_n$, $Y_n$) for new task $n$; Distillation coefficient $\alpha$.
\STATE \textbf{Initialize:}
\STATE \hspace{3pt} Copy teacher Upper Layers to student $h_{\omega} \gets h_{\phi}$
\STATE \textbf{Train:}
\STATE \hspace{3pt} $z \gets$ $f_{\theta}(X_n)$
\STATE \hspace{3pt} $\hat{Y}_{\text{teacher}} \gets h_{\phi}(z)$
\STATE \hspace{3pt}  $\hat{Y}_{\text{student}} \gets h_{\omega}(z)$
\STATE \hspace{3pt}  $\mathcal{L} = \mathcal{L}_{model}(Y_n[c+1:c+k], \hat{Y}_{\text{student}}[c+1:c+k]) + \alpha \cdot \mathcal{L}_{distillation}(\hat{Y}_{\text{student}}[1:c], \hat{Y}_{\text{teacher}}[1:c])$
\STATE \hspace{3pt} Update model parameters: $\omega \gets \mathcal{U}(\mathcal{L}, \omega)$
\STATE \textbf{Return:}
\STATE \hspace{3pt} Model $G(x;\theta;\omega) = h_{\omega}(f_{\theta}(x))$ of task $n$;
\end{algorithmic}
\end{algorithm}

LD, depicted in~\cref{fig:OurApproach}~(right) and outlined in~\cref{alg:LD}, operates by feeding the training samples $X_n$ to a frozen subset of initial layers $f_{\theta}$, shared between the student and teacher, generating representations $z$ in the embedding space. These latent representations are then passed to both the teacher and student Upper Layers ($h_{\phi}$ and $h_{\omega}$, respectively) to compute the distillation and the model loss. The model loss, $\mathcal{L}_{\text{model}}$, only considers new task classes, while the distillation loss, $\mathcal{L}_{\text{model}}$, only considers outputs relative to old classes.
After training, the new model $G(x;\theta;\omega)$ is composed of the frozen Lower Layers $f_{\theta}(x)$ and the student Upper Layers $h_{\omega}$.

The first advantage provided by LD is the reduction of computations for the forward step since the teacher and the student share a common frozen part.
Indeed, as shown in~\cref{fig:OurApproach}~(left), a standard distillation approach requires processing the training images completely through the teacher and student architecture, while for LD, in~\cref{fig:OurApproach}~(right), the logits of the training images produced by $f_{\theta}$ are commonly computed for both the teacher and the student.
Intuitively, increasing the number of frozen layers will improve resource efficiency but it will compromise the model plasticity, as the model has fewer parameters to adapt to the new task. We evaluate this trade-off in~\cref{subsec:freeze}.

The second advantage of LD is that, given that part of the architecture is frozen, the backward function can be performed on a much smaller set of weights.
Intuitively, it is reasonable to freeze the lower layers, as it does not affect significantly the plasticity of future tasks. Indeed, the frozen part belongs to a pre-trained backbone, whose weights do not usually change much during training as they produce general features. The model head, instead, learns how to interpret these features to produce accurate predictions.

Moreover, as the weights of the lower layers are the same for the teacher and student models, they can be stored only once, reducing the memory overhead of standard distillation.    

Although we tested this approach for a specific distillation technique and detection framework, every CL distillation-based approach can be adapted to LD. For this reason, we report implementation details for NanoDet, such as the type of distillation loss, its coefficient, and the training regime in~\cref{sec:experimental_setting}.

\section{Experimental Setting}\label{sec:experimental_setting}
In this section, we outline the criteria we used to evaluate and compare Latent Distillation to other CL methods. We describe the datasets and the considered CL scenarios in~\cref{subsec:dataset} as well as the evaluation metrics in~\cref{subsec:evaluation_metrics}. We then define the CL methods against which we compare our novel approach in~\cref{subsec:considered_methods} for CLOD at the edge.
\subsection{Dataset and Continual Learning Scenario}
\label{subsec:dataset}
We validate our method on PASCAL VOC 2007 and COCO~\cite{everingham2010pascal, lin2014microsoft}, benchmarks commonly used in the CLOD literature~\cite{shmelkov2017incremental,menezes2023continual}.
For VOC we train on the 5k images of the {\em trainval} split and evaluate on the 5k images of the test split.
For COCO, we train on the 80k images of the training set and evaluate on the validation set as in~\cite{shmelkov2017incremental}.
VOC has annotations on 20 object classes, while COCO on 80 classes.
Each task dataset comprises a subset of VOC or COCO containing images with at least one instance of the classes defined in the CL scenario.

We focus on five widely recognized benchmarks in the literature~\cite{menezes2023continual}: 10p10 (read 10-plus-10), 15p5, 19p1, 15p1 on VOC and 40p40 on COCO.
In the 10p10 setup, there are two tasks, each comprising ten classes.
The 15p5 scenario involves a first task with 15 classes and a second task with five.
In the 19p1 configuration, the first task encompasses 19 classes, while the second comprises only one.
We then tackle a more challenging scenario consisting of 6 tasks, denoted as 15p1. Here, the first task includes 15 classes, with each subsequent task introducing a new class.
Lastly, we consider the 40p40 scenario where the first task considers 40 classes and the second one the remaining 40.

The results for the 15p5, 10p10, and 40p40 scenarios are denoted as \textit{"Multiple Classes"}, and for the 19p1 as \textit{"One Class"}. Finally, the 15p1 case is denoted as \textit{"Sequential One Class"}.

\subsection{Evaluation Metrics and Implementation Details}
\label{subsec:evaluation_metrics}
We evaluate our approach using mean Average Precision~(mAP) weighted at different Intersections over Union~(IoU), from 0.5 to 0.95. In particular, we provide three mAP values for each task: \textit{Old}, \textit{New}, and \textit{All}. \textit{Old} reflects the performance on classes encountered in previous tasks, measuring the stability of the model in retaining the previously learned knowledge. \textit{New} denotes the mAP of the current task, indicating the model's plasticity in learning new tasks. Finally, \textit{All} provides an aggregate measure, summarizing the model's overall performance across all the tasks encountered so far.
We also consider Multiply-Accumulate (MAC) computations and the memory overhead required by the CL method.

We use AdamW~\cite{loshchilov2018decoupled} to train Nanodet in all the experiments. The learning rate is set at 0.001 with a 0.05 weight decay.
We use a linear warmup of 500 steps and cosine annealing strategy~\cite{loshchilov2017sgdr} with \(t_{max}=100\).
For each task the network is trained for 100 epochs. As a backbone, we use ShuffleNet V2~\cite{ma2018shufflenet} pre-trained on Image-Net~\cite{krizhevsky2017imagenet}. Moreover, to keep the number of operations low, we take as input images of size 320x320x3.

\subsection{Definition of Compared Methods}
\label{subsec:considered_methods}
The CLOD literature includes solutions that are tailored for specific object detection architecture. Most distillation methods indeed~\cite{hao2019end, peng2020faster, zhou2020lifelong} consider Faster-RCNN, strongly relying on the RPN to get class-agnostic bounding box proposals.
Other approaches~\cite{liu2020incdet, li2019rilod, shieh2020continual} are also dependent on the detection network. Thus, these methods cannot be directly deployed for Nanodet, which is intrinsically different from the models used in the CLOD literature.
Therefore, we evaluate Latent Distillation against classical approaches from the CL literature. Additionally, we implement Selective and Inter-related Distillation (SID)~\cite{peng_sid_2021} because, as pointed out by~\cite{menezes2023continual}, it is the only method designed for an FCOS architecture like Nanodet.

\textbf{Joint Training:} the model is trained on all classes simultaneously; this method represents an upper baseline for CL strategies since it is not affected by catastrophic forgetting.

\textbf{Fine-tuning:} the model is presented sequentially with data from the current task. As no particular technique is used to maintain the model's knowledge, it is considered a lower bound for CL approaches.

\textbf{Replay:} we implement this method~\cite{rolnick2019experience} by creating a buffer at the end of the first task, randomly sampling the task dataset. Then, half of the memory buffer is updated at the end of subsequent tasks by randomly sampling the new task dataset. During training, each batch comprises 50\% images from the task dataset and 50\% randomly sampled from the memory buffer. For each training image the model must be trained on a buffer image, so each training epoch is significantly longer -- an important aspect to consider for applications with constrained resources.
In our experiments, we consider a relatively small replay buffer of 250 images requiring, for NanoDet, 77~MB of additional memory, which can be a challenge for edge devices.

\textbf{Latent Replay:} this approach~\cite{pellegrini2020latent}, is a more resource efficient version of Replay. It achieves this by freezing lower layers of the network, reducing the amount of training computations, and storing the corresponding latent representations of the original task images in the replay buffer, potentially saving memory. We implement this method by freezing the backbone of the model and storing the latent representations of 250 images in the replay buffer. However, Nanodet requires three hidden representations to feed to the feature pyramid, so the buffer requires 81 MB of additional memory, more than the amount needed by standard Replay. For this reason, the efficiency of Latent Replay shows only in terms of faster training times, since the backbone is frozen.
In our experiments, the network is trained fully at task 0, while at subsequent tasks, the backbone is frozen.

\textbf{LwF:} Learning without Forgetting~\cite{li2017learning} is the most well-known technique among the distillation-based approaches. At each training batch, a distillation loss is computed as the Mean Squared Error (MSE) between the logits of the teacher and the student. The distillation loss is then added to the model loss for the task.

\textbf{SID:} Selective and Inter-related Distillation (SID)~\cite{peng_sid_2021} is a distillation technique for CL tailored for FCOS architectures like NanoDet. From this, we replicate the classification head and the intermediate distillation losses as they are the ones that offer the most significant improvement, as shown in the SID paper. For the first one, we mask the teacher and student classification heads to extract only the outputs responsible for predicting the old classes; then we apply a MSE between these filtered logits. For the second one, we consider the outputs of the heads before the split in the classification and regression branches as intermediate distillation. The two distillation losses are then added to the model loss. It is important to note that since NanoDet fuses the ``centerness'' branch in the classification output, the distillation will also carry information about centerness. Additionally, by masking the predictions of new classes, we make the distillation process responsible for remembering only old classes.

\textbf{Latent Distillation:} We implement our approach as described in~\cref{sec:our_approach} using the backbone as the frozen Lower Layers. Then, since NanoDet is an FCOS-style architecture, we add the same type of distillation described for the SID method.
Like Latent Replay, at task 0 we train the whole model, while at subsequent tasks, the backbone is frozen.

All the coefficients of the distillation losses for LwF, SID, and Latent Distillation are set to 1 for the experiments.

\section{Results and Discussion}
\label{sec:results}
In this section, we report the results of LD compared to other CL methods as defined in~\cref{subsec:considered_methods}. These results refer to the Multiple Classes, One Class and Sequential One Class scenarios defined in~\cref{subsec:evaluation_metrics}. 

\begin{table}
\caption{Task performances for the 10p10, 15p5, 19p1 VOC scenarios and the 40p40 COCO scenario. Task values correspond to mAP@[0.50:0.95]. Total Parameters is in million; for Replay approaches the memory buffer is also reported in MB.}\label{tab:VOC+COCOresults}
\centering
\tiny
\begin{tabular}{@{\hspace{5pt}}c@{\hspace{5pt}}c@{\hspace{5pt}}c@{\hspace{5pt}}c@{\hspace{5pt}}c@{\hspace{5pt}}c@{\hspace{5pt}}c@{\hspace{5pt}}c@{\hspace{5pt}}c@{\hspace{5pt}}c@{\hspace{5pt}}c@{\hspace{5pt}}c@{\hspace{5pt}}c@{\hspace{5pt}}c@{\hspace{5pt}}}
\toprule[\heavyrulewidth]
 & & \multicolumn{4}{c}{\textbf{VOC 10 plus 10}} & \multicolumn{4}{c}{\textbf{VOC 15 plus 5}} & \multicolumn{4}{c}{\textbf{VOC 19 plus 1}} \\
\cmidrule(r){3-6} \cmidrule(r){7-10} \cmidrule(r){11-14}
Method & \textbf{Tot. Parameters} & Task 0 & \multicolumn{3}{c}{Task 1} & Task 0 & \multicolumn{3}{c}{Task 1} & Task 0 & \multicolumn{3}{c}{Task 1}\\
\cmidrule(r){4-6} \cmidrule(r){8-10} \cmidrule(r){12-14}
& & & Old & New & \textbf{All} &  & Old & New & \textbf{All} &  & Old & New & \textbf{All} \\
\midrule
Fine-Tuning & 1.2M & \multicolumn{1}{|c}{}& 0 & 31.6 & 15.8 & \multicolumn{1}{|c}{} & 0 & 22.3 & 5.6 & \multicolumn{1}{|c}{} & 2.1 & 23.5 & 3.2 \\
Replay & 1.2M + 77MB& \multicolumn{1}{|c}{} & 7.5 & 24.5 & 16 & \multicolumn{1}{|c}{} & 11.9 & 18.7 & 13.7 & \multicolumn{1}{|c}{} & 15.7 & 27.6 & 16.3\\
Latent Replay & 1.2M + 81MB & \multicolumn{1}{|c}{} & 12.1 & 25 & 18 & \multicolumn{1}{|c}{} & 19.1 & 18.7 & 19.1 & \multicolumn{1}{|c}{} & 16.7 & 26.5 & 17.2\\
LwF & 2.4M & \multicolumn{1}{|c}{32.4} & 13.6 & 29.8 & 21.7 & \multicolumn{1}{|c}{31.9} & 16.4 & 20.4 & 17.4 & \multicolumn{1}{|c}{32.9} & 21.8 & 18.8 & 21.7\\
SID & 2.4M & \multicolumn{1}{|c}{} & 19.1 & 29.8 & \textbf{24.4} & \multicolumn{1}{|c}{} & 19.7 & 19.5 & 19.7 & \multicolumn{1}{|c}{} & 23.2 & 15.7 & 22.8\\
Latent Distillation & \textbf{1.5M} & \multicolumn{1}{|c}{} & 18.9 & 26.4 & 22.7 & \multicolumn{1}{|c}{} & 20.8 & 18.3 & \textbf{20.2} & \multicolumn{1}{|c}{} & 24.3 & 10.9 & \textbf{23.6}\\
\midrule
Joint-Training & & \multicolumn{12}{|c}{30.9}\\
\bottomrule
\end{tabular}
\end{table}
\begin{table}[tb]
\caption{Task performances for the 40p40 COCO scenario. Task values correspond to mAP@[0.50:0.95]. Total Parameters is in million; for Replay approaches the memory buffer is also reported in MB.}\label{tab:COCOresults}
\centering
\tiny
\begin{tabular}{@{\hspace{5pt}}c@{\hspace{5pt}}c@{\hspace{5pt}}c@{\hspace{5pt}}c@{\hspace{5pt}}c@{\hspace{5pt}}c@{\hspace{5pt}}c@{\hspace{5pt}}}
\toprule[\heavyrulewidth]
& & \multicolumn{4}{c}{\textbf{COCO 40 plus 40}} \\
\cmidrule(r){3-6}
Method & \textbf{Tot. Parameters} & Task 0 & \multicolumn{3}{c}{Task 1}\\
\cmidrule(r){3-6}
& & & Old & New & \textbf{All} \\
\midrule
Fine-Tuning & 1.2M & \multicolumn{1}{|c}{} & 0 & 20.1 & 10.1 \\
Replay & 1.2M + 77MB & \multicolumn{1}{|c}{} & 2.2 & 11.8 & 7 \\
Latent Replay & 1.2M + 81MB & \multicolumn{1}{|c}{} & 5.2 & 10.5 & 7.9 \\
LwF & 2.4M & \multicolumn{1}{|c}{25.8} & 3.5 & 16.6 & 10.1 \\
SID & 2.4M & \multicolumn{1}{|c}{} & 15.2 & 16.7 & \textbf{16} \\
Latent Distillation & \textbf{1.5M} & \multicolumn{1}{|c}{} & 15.4 & 11.9 & 13.6 \\
\midrule
Joint-Training & & \multicolumn{4}{|c}{25.1}\\
\bottomrule
\end{tabular}
\end{table}

\subsection{Multiple Classes}
\begin{table}[tb]
\caption{Latent Distillation mAP@[0.50:0.95] results on the 19p1 scenario with progressive backbone unfreezing. Train Parameters refers to the student model, Total parameters considers also those of the teacher. FLOPs overhead refers to the computation required w.r.t distillation approaches like SID or LWF.}
\label{tab:freeze}
\centering
\tiny
\begin{tabular}{@{\hspace{5pt}}c@{\hspace{5pt}}c@{\hspace{5pt}}c@{\hspace{5pt}}c@{\hspace{5pt}}c@{\hspace{5pt}}c@{\hspace{5pt}}c@{\hspace{5pt}}}
\toprule[\heavyrulewidth]
Frozen Layers & Train Param. & Tot. Param. & FLOPs Overhead & Old & New & All \\
\midrule[\heavyrulewidth]
None (SID Distillation) & \multicolumn{1}{|c}{1.2M} & \multicolumn{1}{|c}{2.4M} & \multicolumn{1}{|c}{100\%} & \multicolumn{1}{|c}{23.2} & \textbf{15.7} & 22.8 \\
\midrule
Backbone \textit{except} Stage4ShuffleV2Block[0-1-2-3] & \multicolumn{1}{|c}{899K} & \multicolumn{1}{|c}{2.1M} & \multicolumn{1}{|c}{58\%} & \multicolumn{1}{|c}{23.5} & 13.0 & 22.9 \\
Backbone \textit{except} Stage4ShuffleV2Block[1-2-3] & \multicolumn{1}{|c}{731K} & \multicolumn{1}{|c}{1.93M} & \multicolumn{1}{|c}{52\%} & \multicolumn{1}{|c}{23.7} & 12.1 & 23.1 \\
Backbone \textit{except} Stage4ShuffleV2Block[2-3] & \multicolumn{1}{|c}{620K} & \multicolumn{1}{|c}{1.82M} & \multicolumn{1}{|c}{49\%} & \multicolumn{1}{|c}{23.6} & 12.3 & 23.1 \\
Backbone \textit{except} Stage4ShuffleV2Block[3] & \multicolumn{1}{|c}{598K} & \multicolumn{1}{|c}{1.79M} & \multicolumn{1}{|c}{46\%} & \multicolumn{1}{|c}{24.3} & 10.0 & 23.6 \\
Backbone & \multicolumn{1}{|c}{309K} & \multicolumn{1}{|c}{1.51M} & \multicolumn{1}{|c}{44\%} & \multicolumn{1}{|c}{24.2} & 10.9 & \textbf{23.6} \\
\bottomrule[\heavyrulewidth]
\end{tabular}
\end{table}

In the 10p10, 15p5, and 40p40 scenarios, reported in~\cref{tab:VOC+COCOresults}, SID is confirmed to be the overall better strategy for FCOS-based NanoDet; indeed it is better at maintaining the model knowledge while still being able to learn the new task classes.
Latent Distillation instead, requiring \(74\%\) less additional parameters and \(56\%\) less computation for updating the model with respect to distillation approaches, performs similarly to SID, making our approach suitable for edge applications where small memory footprint and fast training speed are important requirements.
It is interesting to note that Latent Replay benefits from the frozen backbone resulting in less forgetting than Replay. Indeed, as reported in~\cite{tremonti2024empirical}, given a small buffer, better results are obtained if the trainable network is small. However, both Replay approaches are heavily influenced by the limited size of the memory buffer, resulting in high forgetting with bigger task datasets (VOC 10p10 and COCO 40p40).

\subsection{One Class}\label{subsec:freeze}
In the 19p1 scenario, reported in~\cref{tab:VOC+COCOresults}, the full frozen backbone of our approach gives the best stability performance. However, it lacks some learning ability compared to other methods, since it has a smaller number of training parameters and the task dataset includes a limited number of training samples.
It is then useful to compare the learning ability of LD on this task when parts of the backbone are progressively unfrozen.
The results in~\cref{tab:freeze} show that when the model is given more parameters to adapt to the new task, its learning ability improves. However, this implies longer training periods and more parameter overhead as the weights to back-propagate and store increase.
Therefore, choosing how many lower layers should be frozen is a crucial choice based on the compromise between the plasticity of the model and the training/storage efficiency required by the edge application.

We also report in~\cref{fig:results}~(right) the overall mAP in this task vs the total model parameters that need to be stored by the CL strategy. Our method achieves the best performance when considering these two metrics.
It is important to notice that total parameters refers only to the model parameters needed by the CL strategy; for the Replay approaches, it does not take into account that they require an additional memory buffer, which can be a significant problem for some edge applications, as mentioned in~\cref{subsec:considered_methods}.

\subsection{Sequential One Class}\label{subsec:15p1}
\begin{table}[tb]
\caption{Comparison of Task Performances for the 15p1 (Sequential One Class) VOC scenario. Values are defined as in ~\cref{tab:VOC+COCOresults}.}
\label{tab:15p1}
\centering
\tiny
\begin{tabular}{@{\hspace{5pt}}c@{\hspace{5pt}}c@{\hspace{5pt}}c@{\hspace{5pt}}c@{\hspace{5pt}}c@{\hspace{5pt}}c@{\hspace{5pt}}c@{\hspace{5pt}}c@{\hspace{5pt}}c@{\hspace{5pt}}c@{\hspace{5pt}}c@{\hspace{5pt}}c@{\hspace{5pt}}c@{\hspace{5pt}}c@{\hspace{5pt}}c@{\hspace{5pt}}c@{\hspace{5pt}}}
\toprule[\heavyrulewidth]
Method & \multicolumn{3}{c}{Task 1} & \multicolumn{3}{c}{Task 2} & \multicolumn{3}{c}{Task 3} & \multicolumn{3}{c}{Task 4} & \multicolumn{3}{c}{Task 5} \\
\cmidrule(r){2-4} \cmidrule(r){5-7} \cmidrule(r){8-10} \cmidrule(r){11-13} \cmidrule(r){14-16}
& Old & New & \textbf{All} & Old & New & \textbf{All} & Old & New & \textbf{All} & Old & New & \textbf{All} & Old & New & \textbf{All} \\
\midrule[\heavyrulewidth]
Fine-Tuning & \multicolumn{1}{|c}{7.5} & 9.1 & 7.2 & \multicolumn{1}{|c}{1.6} & 4.3 & 1.77 & \multicolumn{1}{|c}{0} & 12.8 & 0.7 & \multicolumn{1}{|c}{0} & 8.4 & 0.4 & \multicolumn{1}{|c}{0} & 12.8 & 0.6 \\
Replay & \multicolumn{1}{|c}{21} & 8.7 & 20.3 & \multicolumn{1}{|c}{18.2} & 7.1 & \textbf{17.5} & \multicolumn{1}{|c}{9.9} & 12.1 & 10.1 & \multicolumn{1}{|c}{6.6} & 25.6 & 7.7 & \multicolumn{1}{|c}{4.1} & 21.5 & 5 \\
Latent Replay & \multicolumn{1}{|c}{22.9} & 7.9 & 20.5 & \multicolumn{1}{|c}{19.1} & 10.9 & \textbf{17.5} & \multicolumn{1}{|c}{12.8} & 18 & 12.4 & \multicolumn{1}{|c}{9.7} & 19 & 9.7 & \multicolumn{1}{|c}{5.3} & 21.3 & 6.1 \\
LwF & \multicolumn{1}{|c}{20.2} & 6.2 & 19.4 & \multicolumn{1}{|c}{8} & 2.8 & 7.8 & \multicolumn{1}{|c}{1.3} & 9.6 & 1.8 & \multicolumn{1}{|c}{0.5} & 15 & 1.3 & \multicolumn{1}{|c}{0.3} & 19.6 & 1.3 \\ 
SID & \multicolumn{1}{|c}{21.3} & 6.4 & 20.4 & \multicolumn{1}{|c}{15.1} & 3.2 & 14.4 & \multicolumn{1}{|c}{16.2} & 3.9 & \textbf{15.6} & \multicolumn{1}{|c}{10.8} & 4 & 10.4 & \multicolumn{1}{|c}{10.1} & 10.1 & \textbf{10.1} \\
Latent Distillation & \multicolumn{1}{|c}{23.9} & 5.7 & \textbf{22.8} & \multicolumn{1}{|c}{18.1} & 4.8 & 17.3 & \multicolumn{1}{|c}{15.6} & 1.3 & 14.8 & \multicolumn{1}{|c}{11.5} & 1.3 & \textbf{11} & \multicolumn{1}{|c}{10} & 2.1 & 9.6 \\
\midrule
Joint-Training & \multicolumn{14}{|c}{30.88} \\
\bottomrule[\heavyrulewidth]
\end{tabular}
\end{table}

\begin{figure}[tb]
\begin{tabular}{cc}
\includegraphics[width=0.5\linewidth]{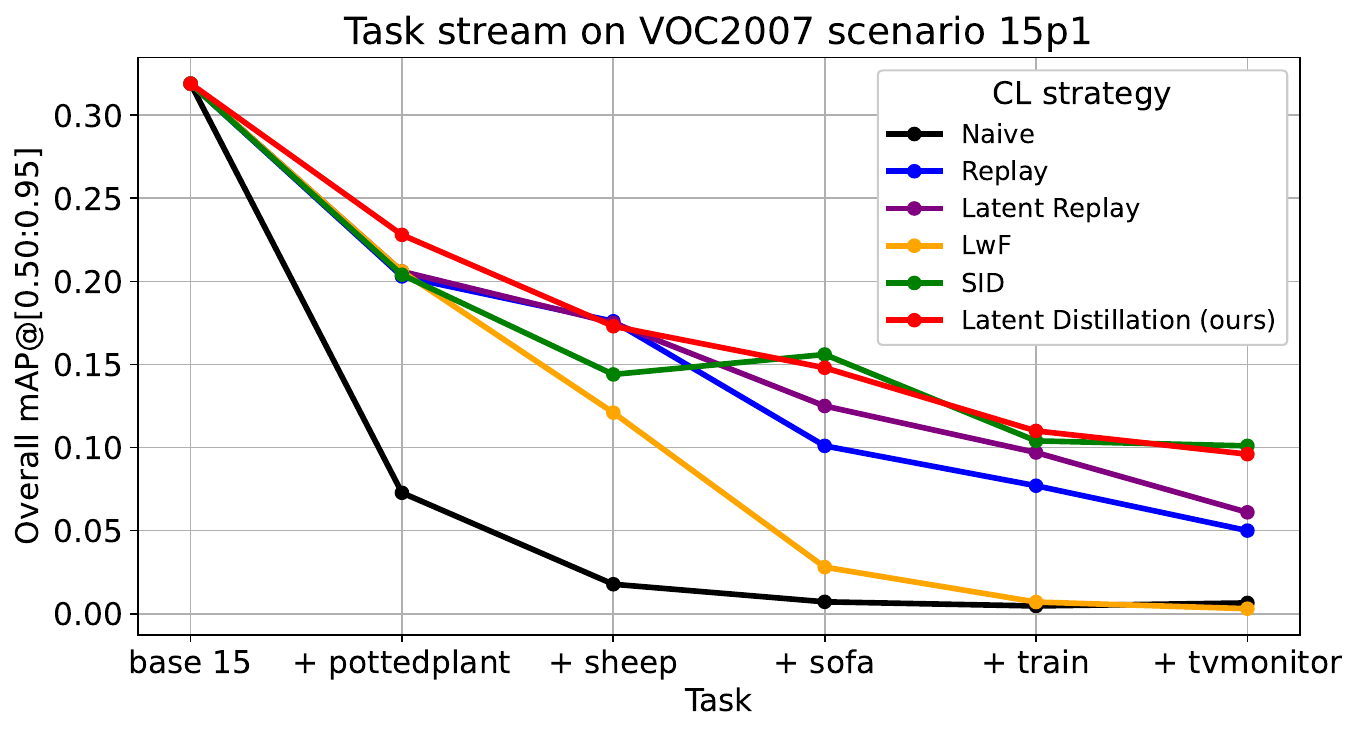}&
\includegraphics[width=0.5\linewidth]{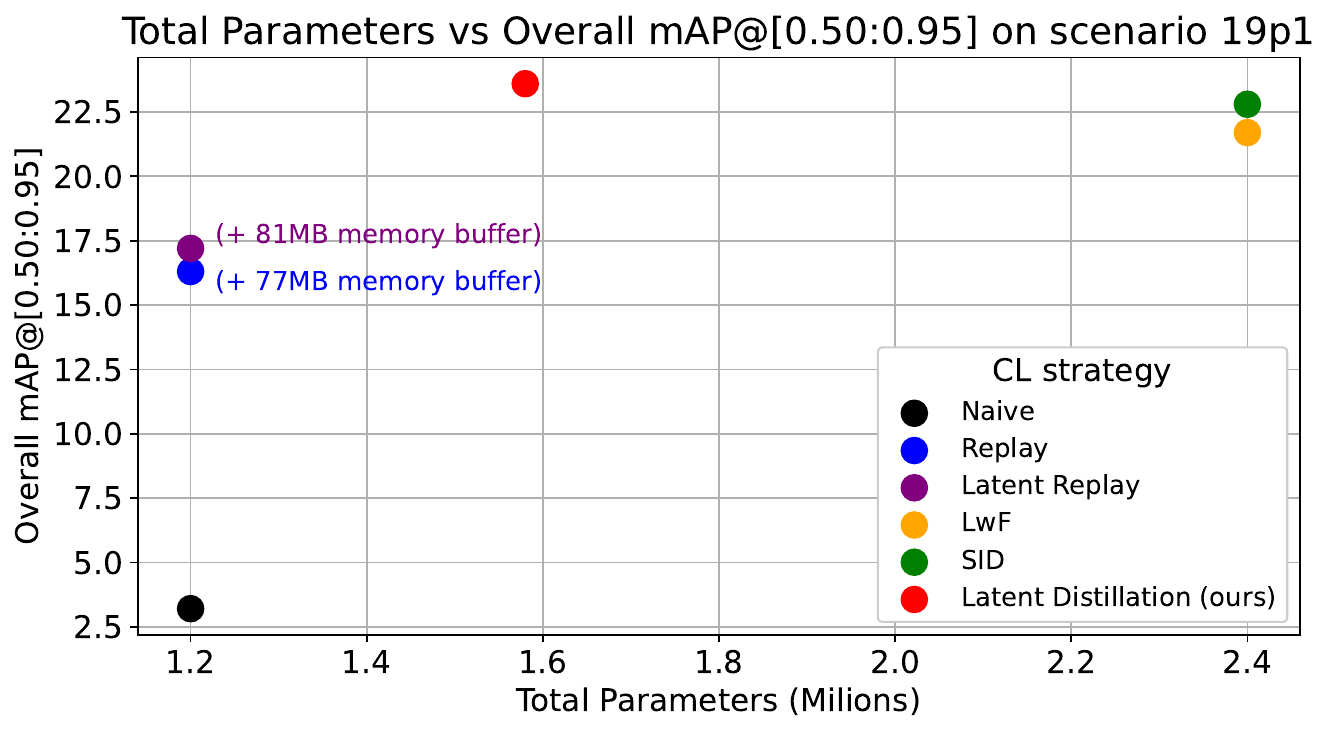}\\
\end{tabular}
\caption{\textbf{Left:} Results on the Sequential One Class (15p1) scenario. \textbf{Right:} Overall mAP vs total parameters required by the CL method on the 19p1 scenario. Our method shows the best overall performances when considering the two metrics.}
\label{fig:results}
\end{figure}

We finally consider the 15p1 scenario where the model must learn one class in sequential updates, starting from a base model trained on the first 15 classes.
The results reported in~\cref{tab:15p1} confirm SID and LD are the best compromise between stability and plasticity. Indeed, both strategies provide the best results in terms of forgetting, while also being able to learn new classes.
In some scenarios, the entire frozen backbone of our approach results in less forgetting than all the other methods. 
However, LD shows limited plasticity for some task datasets: this is an expected behavior, as already discussed in~\cref{subsec:freeze} and showed in~\cref{tab:freeze}. If the constraints imposed by the edge application permit it, choosing to freeze a smaller part of the network would lead to better results.
Moreover, it can be noticed that, while LwF performs relatively well on the first task, the noise caused by the regression outputs eventually leads to catastrophic forgetting during the subsequent model updates, as reported in~\cite{peng_sid_2021}.
The results of this experiment are also depicted in~\cref{fig:results}.

\section{Conclusion and Future Works}
\label{sec:conclusions_future_works}
We investigated using an efficient NanoDet detector for CLOD applications at the edge. We evaluated this detector on several CL scenarios, proving that it can learn incrementally new classes over time while retaining knowledge of previously learned classes. We then proposed a new Latent Distillation approach, which avoids the storage of multiple copies of the same architecture concurrently during training, as demanded by previous distillation-based methods. 
Our solution reduces the distillation memory overhead by 74\% and computational burden by 56\%, with a minor performance decrease in most scenarios.

In the future, we want to assess the generalizability of Latent Distillation in other CIL and DIL scenarios using different models and backbones.
We also plan to implement this framework in a tinyML device.
Additionally, we plan to expand the study to other recent and efficient detectors suitable for edge deployment and evaluate CLOD in different CL real-world scenarios with tiny autonomous robots.

\par\vfill\par
%
%
\bibliographystyle{splncs04}
\bibliography{main}
\end{document}